\documentclass[letterpaper, 10 pt, conference]{ieeeconf}

\IEEEoverridecommandlockouts
\usepackage{times}
\usepackage{amsmath,amssymb}
\usepackage{graphicx}
\usepackage{xcolor}
\usepackage{booktabs}
\usepackage{multirow}
\usepackage{url}
\usepackage{cite}
\usepackage[hidelinks]{hyperref}
\usepackage{xurl}

\usepackage{algorithm}
\usepackage{algpseudocode}
\usepackage{xspace}
\usepackage{listings}
\usepackage{adjustbox}
\usepackage{wrapfig}
\usepackage{comment}

\usepackage{booktabs}
\usepackage{siunitx}     % 数字对齐（可选）
\usepackage{threeparttable} % 表注/脚注
\usepackage{caption}
\usepackage{dsfont}

\usepackage{wrapfig}

\newcommand{\method}{{\it SPEAR}\xspace}

\title{\LARGE \bf
\method: Selective Preference Elicitation with Adaptive Rewards for VLM-Guided Robotic Manipulation
}

\author{
Zewen Wu$^{1}$, Ruizhe Liu$^{1}$, Pei Zhou$^{1}$,
Feng Chen$^{1}$, and Yanchao Yang$^{1,2,*}$%
\thanks{$^{1}$HKU Musketeers Foundation Institute
of Data Science, The University of Hong Kong,
Hong Kong SAR, China.
{\tt\small zwwu@connect.hku.hk, yanchaoy@hku.hk}}%
\thanks{$^{2}$Department of Electrical and Electronic Engineering, 
The University of Hong Kong, Hong Kong SAR, China.}
\thanks{$^{*}$Corresponding author.}%
}

\begin{document}
\maketitle
\thispagestyle{empty}
\pagestyle{empty}

% =======================
% Abstract
% =======================
\begin{abstract}
% TODO: Replace with your abstract.
% NOTE: Do NOT wrap with \begin{abstract}...\end{abstract}; root.tex already does.
Recent Vision-Language Models (VLMs) 
enable automated preference labeling 
for robotic manipulation,
yet their deployment 
remains impractical since each query incurs
substantial inference cost,
while VLMs frequently abstain on visually challenging comparisons.
We propose \method, 
a framework that transforms 
VLMs into efficient preference oracles 
via two synergistic mechanisms. 
First, we design \emph{distinguishability-aware query selection} to reduce abstention rates by targeting only informative and distinguishable comparisons.
Second, we deploy an \emph{adaptive reward transition} that bootstraps exploration with VLM-synthesized guidance before preference-based refinement.
Experiments on the manipulation benchmarks demonstrate that \method\ achieves 88.9\% average success, outperforming baselines by 23.3 percentage points, while significantly reducing abstention and wall-clock training time on contact-rich tasks.
% Ablations confirm that both mechanisms are essential.
\method\ learns three Mobile ALOHA manipulation tasks fully online on real hardware, demonstrating end-to-end real-world applicability.
By addressing query efficiency 
and exploration challenges, \method~establishes VLMs 
as practical preference oracles 
for scalable robot learning without manual reward engineering. Project page:
{\urlstyle{same}\url{https://zevinwu.github.io/spear-project}}.
% Diagnostics across multiple open-source VLM backbones show that higher visual distinguishability simultaneously lowers abstention and improves preference accuracy, supporting abstention-aware acquisition as a core design principle.

\end{abstract}

% =======================
% Main content
% =======================
\section{Introduction}
\label{sec:introduction}

Reinforcement learning (RL)
offers a principled way to
acquire robotic manipulation skills in demonstration-scarce settings, yet practical deployment remains constrained by reward specification. 
Preference-based Reinforcement Learning (PbRL) 
addresses this by learning from pairwise trajectory comparisons rather than numerical rewards~\cite{christiano2017deep,wirth2017survey}. 
However, the scalability of PbRL is constrained by the need for frequent human-in-the-loop preference queries.
Recent advances in Vision-Language Models (VLMs) 
present an appealing solution: 
modern VLMs can naturally serve as automated preference oracles~\cite{wang2024rl} 
due to their encoding of extensive world knowledge and robust visual-language grounding.
This enables automated robot learning from online interaction and language goals, 
without relying on demonstrations, manually designed rewards, or ongoing human labeling.

However, 
VLMs introduce a notable computational constraint -- 
each preference query incurs a significant inference cost.
This makes query efficiency critical,
yet existing VLM-guided PbRL methods waste considerable budgets due to two fundamental inefficiencies.
{\it First,} 
VLMs are frequently incapable of providing valid preferences for visually similar or indistinguishable trajectories, i.e., returning abstention.
% Moreover, reliable deployment requires conservative prompting to mitigate systematic positional bias and cross-image attribution errors, which further induces abstention on ambiguous comparisons.
Naive query selection
could misuse a significant portion
of expensive queries
on comparisons that yield zero learning signal.
Across backbones, higher visual distinguishability leads to both lower abstention and higher preference accuracy, motivating abstention-aware acquisition as a core design target.
{\it Second,} 
meaningful preferences 
require behaviorally task-related trajectories, 
yet initial random policies produce only aimless motions. 
This creates a bootstrapping paradox: 
generating task-related behaviors requires rewards that need to be learned from comparing task-related behaviors. 
{\it Together,} these dilemmas make VLM-guided preference learning inefficient for automated learning.

We present \method (\underline{\textbf{S}}elective \underline{\textbf{P}}reference \underline{\textbf{E}}licitation with \underline{\textbf{A}}daptive \underline{\textbf{R}}ewards),
a framework that addresses both challenges with two synergistic mechanisms designed specifically for VLM-based preference learning under computational constraints.
{\it To overcome} the query efficiency challenge, 
we develop \textbf{Distinguishability-Aware Query Selection} 
that predicts VLM response likelihood before querying. 
Our approach combines global semantic embeddings with local spatial patterns, 
% capturing both high-level task understanding and fine-grained manipulation details. 
and further trains a calibrated distinguishability prediction model. 
By weighting information gain with predicted distinguishability, 
we ensure computational resources target only answerable and informative comparisons, while enhancing learning efficiency.
{\it To resolve} 
the exploration bootstrapping challenge, 
we introduce \textbf{Adaptive Reward Transition} that leverages VLM-synthesized rewards to jumpstart exploration before transitioning to refined preference-based learning. 
While VLM-synthesized rewards lack precision for task completion, 
they successfully guide initial exploration toward semantically task-relevant regions.
Combined with relay exploration that maintains distributional coverage, 
this mechanism ensures continuous generation of diverse yet task-relevant trajectories that are necessary for meaningful preference comparisons.

We evaluate our approach on six diverse manipulation tasks from Meta-World~\cite{yu2020meta}.
\method achieves an 88.9\% average success rate, surpassing the strongest VLM-based baseline by 23.3 percentage points, with substantially fewer wasted oracle queries and reduced wall-clock training time.
Beyond simulation, we further validate \method on a real Mobile ALOHA robot by running the full online pipeline directly in the real world, demonstrating VLM-guided preference learning can bootstrap manipulation skills from scratch under realistic sensing and actuation constraints.
The ablation studies confirm that both mechanisms are essential for efficient online PbRL. Our contributions are threefold:
\begin{itemize}
    \item {\it First,} we propose a distinguishability-aware query selection mechanism that predicts VLM answerability through multi-scale visual features and online-adapted modeling, effectively reducing wasted queries. 
    \item {\it Second,} we introduce an adaptive reward transition strategy 
    that bootstraps exploration with VLM-synthesized reward before transitioning to preference-based learning, ensuring the collection of informative trajectories. 
    \item {\it Third,} we demonstrate substantial improvements across diverse manipulation tasks and on a real robot,
    validating that addressing query efficiency and exploration bootstrapping enables practical VLM-guided preference learning under realistic computational constraints.
\end{itemize}

\section{Related Work}
\label{sec:related_work}

\subsection{VLM-guided Robot Policy Learning.} 
The difficulties in hand-crafting 
dense reward functions 
for robotic control 
have motivated a new paradigm leveraging recent % [EDIT] has -> have (grammar)
large-scale models for reward specification or generation \cite{rocamonde2023vision,xie2023text2reward,wang2024rl}. 
A typical current approach 
utilizes Vision-Language Models (VLMs) 
% [DEL] prune citation dump; keep only representative backbones/surveys
\cite{li2023blip,zhu2025internvl3,yang2025qwen3,li2025survey,wang2025internvl3.5} 
% driess2023palm
as direct, zero-shot reward models (VLM-RMs) \cite{rocamonde2023vision}  
by calculating the cosine similarity 
between the visual and the goal text embedding. 
This strategy has also been explored 
in works using CLIP-style models \cite{cui2022can,mahmoudieh2022zero,sontakke2023roboclip}. 
Other works, exemplified by Text2Reward \cite{xie2023text2reward}, 
use Large Language Models (LLMs) to generate executable dense reward programs grounded in environment APIs \cite{yu2023language,ma2023eureka}. 
An alternative paradigm, 
which our work adopts, 
avoids direct reward specification. 
{\it Instead,} 
it leverages VLMs as automated preference oracles within a Preference-based Reinforcement Learning (PbRL) framework. 
For instance, 
RL-VLM-F \cite{wang2024rl} queries a VLM to provide preference labels for pairs of image observations, and then learns a reward function from this feedback. 
% This aligns with a broader trend of integrating foundation models into robotic control \cite{brohan2022rt, jiang2022vima}. 
% [EDIT] align with our main storyline (abstention/answerability + costly queries), avoid over-claiming "label reliability"
Our approach builds upon this VLM-driven preference-based paradigm, 
focusing on the query efficiency and effective preference elicitation under costly, abstaining VLM oracles. 
% [ADD][optional, 1 sentence positioning; keep short, no expansion]
% Our focus is complementary to data-rich BC/VLA pipelines, targeting demo-scarce online learning settings.

\subsection{Preference-based Reinforcement Learning.}
PbRL\cite{christiano2017deep,wirth2017survey,xu2020preference,lee2021b,lee2021pebble,park2022surf,liu2022meta,liu2023efficient} 
learns an implicit reward function from pairwise comparisons provided by an expert, often based on the Bradley-Terry model \cite{Bradley1952Rank}. 
Traditionally, these comparisons are sourced from human experts \cite{christiano2017deep, ibarz2018reward, leike2018scalable}. 
{\it However,} this process is notoriously costly and slow, creating an annotation bottleneck that hinders scalability. {\it Meanwhile,} 
many existing PbRL systems and benchmarks assume access to an idealized scripted teacher. 
% [EDIT] soften tone; clarify as upper bound / benchmark assumption
This oracle computes preferences by using privileged access to the ground-truth environment reward signals \cite{lee2021pebble}, an approach that constitutes an unrealistic training regime and contradicts the PbRL desideratum. 
The concurrent advancement of VLMs offers a promising direction, positioning VLMs as natural candidates to serve as automated preference oracles \cite{wang2024rl}. 
Moreover, 
AI feedback (RLAIF) has also been explored for aligning models \cite{bai2022constitutional,klissarov2023motif}. 
While this paradigm shift addresses the expertise bottleneck, 
it introduces new, open problems: 
how to query these powerful but fallible and non-negligibly costly VLM oracles efficiently and effectively remains challenging.

\section{Preliminaries}
\label{sec:preliminaries}

% [KEEP] Motivation for PbRL with VLMs (light touch edits for consistency).
% To automate learning of manipulation policies with vision-language models (VLMs),
% we choose to work with the preference-based reinforcement learning (PbRL) framework,
% as it naturally aligns with VLMs' comparative reasoning capabilities---rather than requiring VLMs to output precise numerical rewards for individual states.
% Therefore, we can maximally leverage VLMs' contrastive visual-language understanding
% and their ability to reason about \emph{relative} quality rather than absolute values.

\begin{figure*}
  \centering
  \includegraphics[width=17cm]{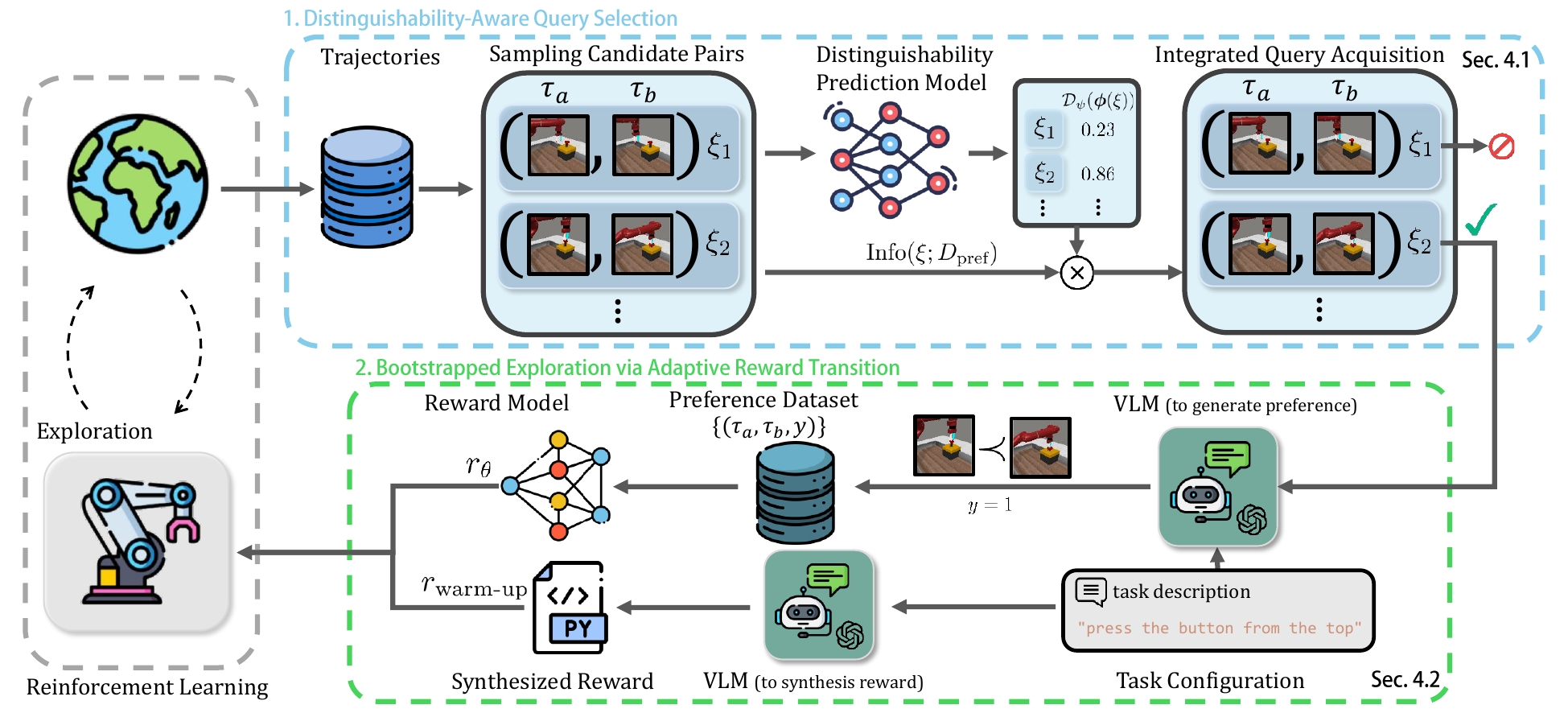}
  \caption{
    \textbf{Overview of \method.} 
    % Our framework integrates two key mechanisms: 
    \textbf{(1) Distinguishability-aware query selection} (top, blue) predicts VLM response likelihood via multi-scale features and weights information gain accordingly, selecting only answerable and informative trajectory pairs for querying;
    \textbf{(2) Adaptive reward transition} (bottom, green) bootstraps exploration using VLM-synthesized warm-up rewards before transitioning to refined preference-based learning. 
    Together, these mechanisms enable efficient online learning by ensuring both high-quality query selection and sufficient behavioral diversity for meaningful preference elicitation.
  }
  \label{overviewfig}
\end{figure*}

% [EDIT] MDP -> POMDP, and make image observation explicit (policy is image-based).
% \noindent\textbf{Problem Formulation.} 
\subsubsection{Problem Formulation} 
Formally, we model robotic manipulation with image observations as a Partially Observable Markov Decision Process (POMDP)
$\mathcal{M} = (\mathcal{S}, \mathcal{A}, \mathcal{P}, \gamma, \mathcal{O}, \Omega)$,
where $\mathcal{S}$ denotes the (latent) state space,
$\mathcal{A}$ the action space,
$\mathcal{P}: \mathcal{S} \times \mathcal{A} \rightarrow \Delta(\mathcal{S})$ the transition dynamics,
$\gamma \in [0,1)$ the discount factor,
$\mathcal{O}$ the observation space, and
$\Omega(o \mid s)$ the observation model.
In our setting, observations are images, denoted by $o_t = \mathcal{I}(s_t) \in \mathcal{O}$,
and the policy acts on observations $\pi: \mathcal{O} \rightarrow \Delta(\mathcal{A})$.
Unlike standard RL, we do not assume access 
to a reward function 
$r: \mathcal{S} \times \mathcal{A} \rightarrow \mathbb{R}$. 
Instead, we aim to learn 
policies through the preference feedback of the VLM while training a 
surrogate reward model. 

% [ADD] Define the VLM preference oracle interface and abstention semantics (no long prompt).
% \noindent\textbf{VLM preference oracle.}
\subsubsection{VLM Preference Acquisition} 
Let $\tau = \{o_t\}_{t=1}^{H}$ 
denote an observation sub-trajectory of horizon $H$. Given a task description $T$ and two observation sub-trajectories 
$\tau^a, \tau^b$, we query a VLM oracle to return a preference label $y \in \{0, 1, \varnothing\}$, 
where $y = 0$ indicates $\tau^a \succ \tau^b$ (sub-trajectory $a$ is preferred over $b$), 
$y = 1$ indicates $\tau^b \succ \tau^a$, 
and $y = \varnothing$ denotes abstention (no clear progress difference or insufficient evidence). 
% The training objective 
% is a policy $\pi: \mathcal{S} \rightarrow \Delta(\mathcal{A})$ 
% that maximizes the expected return with the surrogate reward function induced by preference feedback as in the following.

% [KEEP] Bradley-Terry preference modeling, with minimal edits to use observations.
% \noindent\textbf{Reward by preference modeling.}
\subsubsection{Reward Modeling}
To obtain a surrogate reward function, we adopt Bradley--Terry modeling~\cite{Bradley1952Rank}.
We instantiate a neural reward model $r_\theta: o_t \rightarrow \mathbb{R}$ with parameters $\theta \in \Theta$
and model the likelihood of preference as:
\begin{equation}
P_\theta[\tau^b \succ \tau^a] =
\sigma\left(
\sum_{t} r_\theta(o_t^b) -
\sum_{t} r_\theta(o_t^a)
\right),
\label{eq:bradley-terry}
\end{equation}
where $\sigma(z) = (1 + e^{-z})^{-1}$ is the sigmoid function.
Given a dataset
$D_\text{pref} =
\{(\tau^{a,n}, \tau^{b,n}, y^n)\}_{n=1}^{N}$
of labeled sub-trajectory pairs (with abstentions excluded),
we optimize $\theta$ by maximizing the log-likelihood:
\begin{equation}
\label{eq:preference-likelihood}
\begin{split}
\mathcal{L}(\theta; D_\text{pref}) =
\sum_{n=1}^{N} \big(
y^n \log P_\theta[\tau^{b,n} \succ \tau^{a,n}] + \\
(1-y^n) \log P_\theta[\tau^{a,n} \succ \tau^{b,n}]
\big).
\end{split}
\end{equation}

% [KEEP] General PbRL loop (light edits only if needed later).
% \noindent\textbf{The general PbRL learning framework}
\subsubsection{The General PbRL Learning Framework}
The whole training loop operates by iterating through three steps:
% \begin{enumerate} % [leftmargin=*, topsep=0pt, itemsep=2pt]
\begin{itemize}
    \item \textit{Exploration:}
    The current policy $\pi$ trained with the most recent $r_\theta$
    interacts with the environment while collecting candidate trajectories.
    \item \textit{Query Selection and Preference Prediction:}
    A sampling mechanism selects a set of sub-trajectory pairs
    $\{(\tau^{a,n}, \tau^{b,n})\}_{n=1}^{N}$
    for preference elicitation with a VLM oracle, forming
    $D_\text{pref} = \{(\tau^{a,n}, \tau^{b,n}, y^n)\}_{n=1}^{N}$.
    Since each VLM query incurs non-negligible inference cost and may abstain,
    the sampling mechanism critically determines query efficiency.
    \item \textit{Reward Learning:}
    The preference dataset is augmented
    $D_\text{pref}' \leftarrow D_\text{pref}' \cup D_\text{pref}$,
    and the reward model $r_\theta$ is updated via Eq.~\ref{eq:preference-likelihood}.
\end{itemize}
% \end{enumerate}

The above steps 
are iterated until the policy performance converges 
or a pre-defined budget is reached.

% =========================================================
% SPEAR Section Intro (compressed for IROS)
% =========================================================
\section{\method: Selective Preference Elicitation with Adaptive Rewards}
\label{sec:method}

% [EDIT] Compressed intro to avoid repeating Introduction.
Deploying VLMs as preference oracles in online PbRL is bottlenecked by two practical issues:
(i) expensive queries frequently yield abstentions on visually indistinguishable comparisons, and
(ii) early exploration lacks task progress, producing uninformative pairs.
\method addresses both via
(a) \textbf{Distinguishability-Aware Query Selection} (Sec.~\ref{sec:method:Preference-Acquisition}), which spends queries only on answerable, informative pairs, and
(b) \textbf{Adaptive Reward Transition} (Sec.~\ref{sec:exploration}), which bootstraps exploration with a coarse warm-up reward before switching to preference-based refinement.

% =========================================================
% Block A: Distinguishability-aware Query Selection (IROS)
% =========================================================

\subsection{Distinguishability-Aware Query Selection}
\label{sec:method:Preference-Acquisition}

% [EDIT] A0: One motivating paragraph (avoid repeating Introduction).
Classical acquisition functions (e.g., BALD~\cite{houlsby2011bayesian}) assume an ideal oracle that always provides effective labels.
However, VLM oracles often abstain on trajectory pairs lacking clear visual differences, making expensive queries yield zero learning signal.
Moreover, to enhance the accuracy of VLM preferences, we adopt a decomposed and conservative prompting protocol that mitigates positional bias and cross-image attribution errors. This makes abstention a systematic and unavoidable outcome for low-distinctiveness pairs.
Empirically, increasing visual distinctiveness simultaneously reduces abstention and improves label accuracy (see Sec.~\ref{sec:oracle_diagnostics}). Therefore,  prioritizing distinguishable pairs effectively avoids both wasted queries and low-quality supervision.

We address the issue of avoiding queries that may lead to abstention and unreliable labels through \emph{distinguishability-aware acquisition}:
a mechanism that predicts VLM response likelihood via multi-scale visual features (Sec.~\ref{sec:method:features}),
maintains a distinguishability model (Sec.~\ref{sec:method:distinguish_model}),
and integrates these predictions with information gain to select only answerable, informative pairs (Sec.~\ref{sec:method:acquisition}).

\subsubsection{Multi-Scale Visual Feature Extraction}
\label{sec:method:features}

% [KEEP] (light touch edits only): original rationale.
Predicting the distinguishability between two sub-trajectories conditioned on the VLM
requires understanding both the task semantics and fine-grained manipulation progress.
We therefore construct features that capture
high-level semantic differences (which VLMs excel at)
and subtle spatial variations (which VLMs may struggle with):

% \noindent\textbf{Global semantic features.}
We extract global embeddings $\mathbf{e}^a, \mathbf{e}^b \in \mathbb{R}^c$ from the VLM's vision encoder
for the sub-trajectories $\tau^a, \tau^b$, then compute relational features:
\begin{equation}
\boldsymbol{\phi}_{\text{rel}}(\xi) =
[|\mathbf{e}^a - \mathbf{e}^b|;\ \mathbf{e}^a \odot \mathbf{e}^b;\ \cos(\mathbf{e}^a, \mathbf{e}^b)]
\in \mathbb{R}^{2c+1},
\end{equation}
where $\xi = (\tau^a,\tau^b)$ denotes a sub-trajectory pair.
These features capture complementary similarity measures:
absolute differences indicate semantic distance,
element-wise products capture feature co-activation,
and cosine similarity measures directional alignment.

% \noindent\textbf{Local spatial features.}
Global embeddings may miss subtle spatial differences crucial for manipulation tasks,
e.g., whether a gripper is slightly open versus closed.
We employ cross-attention between spatial token grids $\mathbf{T}^a, \mathbf{T}^b \in \mathbb{R}^{N \times D}$
to identify such discrepancies.
After projecting spatial tokens into a lower-dimensional space $\mathbf{T}^a_{\text{p}}, \mathbf{T}^b_{\text{p}}\in\mathbb{R}^{N \times d}$,
bidirectional cross-attention produces aligned representations $\tilde{\mathbf{T}}^a, \tilde{\mathbf{T}}^b$.
We compute statistics capturing alignment quality:
the mean and maximum of $|\mathbf{T}^a_{\text{p}} - \tilde{\mathbf{T}}^a|$ (misalignment strength)
and $\mathbf{T}^a_{\text{p}} \odot \tilde{\mathbf{T}}^a$ (alignment confidence),
yielding $\boldsymbol{\phi}_{\text{loc}}(\xi) \in \mathbb{R}^{8d}$.
The combined feature
$\boldsymbol{\phi}(\xi) = [\boldsymbol{\phi}_{\text{rel}}(\xi);\boldsymbol{\phi}_{\text{loc}}(\xi)]$
provides a comprehensive signature of visual distinguishability across semantic and spatial scales.

\subsubsection{Distinguishability Prediction Model}
\label{sec:method:distinguish_model}

% [KEEP + EDIT] Keep core, add explicit periodic sliding-window recalibration.
Given features $\boldsymbol{\phi}(\xi)$, we train a calibrated predictor
$\mathcal{D}_{\psi}: \mathbb{R}^{2c+8d+1} \rightarrow [0, 1]$
that estimates the probability that the VLM returns a non-abstention label (versus abstention).
Crucially, this predictor must adapt online as the policy explores new regions where VLM behavior may differ.

% \noindent\textbf{Calibrated prediction.}
We instantiate $\mathcal{D}_{\psi}$ as a lightweight MLP trained on accumulated VLM responses
$\{(\boldsymbol{\phi}(\xi_i), \mathds{1}[y_i \neq \varnothing])\}$.
Since raw neural predictors are often miscalibrated~\cite{guo2017calibration},
we apply temperature scaling with parameter $T$.
In practice, we periodically recalibrate $T$ on a sliding window of recent queries to maintain reliable probabilities under distribution shift.
The distinguishability model continuously learns from new VLM responses throughout training.
% \noindent\textbf{Online adaptation.}
% The distinguishability model continuously learns from new VLM responses throughout training.
% After each query batch, we augment the training data and update both $\mathcal{D}_{\psi}$
% (and the local cross-attention module for feature extraction).
% This adaptation is essential as the policy evolves:
% early random trajectories may appear uniformly similar to the VLM,
% while later skilled behaviors reveal subtle differences the VLM can distinguish.

\subsubsection{Integrated Query Acquisition}
\label{sec:method:acquisition}

% [KEEP + EDIT] Keep equation; avoid posterior details; add one sentence: estimated via standard PbRL approximations.
We combine distinguishability prediction with information-theoretic acquisition to maximize learning efficiency per VLM query.
% \noindent\textbf{Distinguishability-weighted information gain.}
Standard BALD scores a pair by expected information gain to the reward model:
\begin{equation}
\text{Info}(\xi; D_\text{pref}) =
\mathbb{H}[y \mid \xi] -
\mathbb{E}_{p(\theta \mid D_\text{pref})}[\mathbb{H}[y \mid \xi, \theta]],
\end{equation}
where $\xi = (\tau^a, \tau^b)$ and $D_\text{pref}$ is the current preference dataset.
We estimate $\text{Info}(\xi; D_\text{pref})$ following standard PbRL practice \cite{lee2021pebble},
and note that information gain becomes meaningless if the VLM abstains.
We therefore weight by predicted distinguishability:
\begin{equation}
\text{Info}_{\text{\method}}(\xi) =
\mathcal{D}_{\psi}(\boldsymbol{\phi}(\xi)) \cdot \text{Info}(\xi; D_\text{pref}).
\label{eq:vlm-aware-score}
\end{equation}
This product balances two objectives:
selecting pairs that resolve reward-model uncertainty (high $\text{Info}$)
while avoiding pairs the VLM cannot distinguish (low $\mathcal{D}_{\psi}$).

% \noindent\textbf{Diversified batch selection.}
Since many similar pairs may achieve high scores, directly selecting top-ranked pairs yields redundant queries.
We apply facility-location optimization to select diverse batches that maximize coverage while maintaining high scores.
Concretely, we perform diversification in the VLM embedding space $\boldsymbol{\phi}(\xi)$ with cosine distance,
so selected queries span diverse progress states instead of near-duplicates.

% [ADD] Box-1: Query protocol (short; do NOT paste full prompts).

\subsubsection{VLM Query Protocol}
Given task $T$ and two observation sub-trajectories $\tau^a,\tau^b$, our initial attempts start from the prompt template of RL-VLM-F~\cite{wang2024rl}, but we observe two systematic failure modes that corrupt preference learning:
(i) \emph{positional bias}, where the predicted preference correlates with presentation order rather than content, e.g., disproportionately favoring the latter candidate, and
(ii) \emph{cross-image attribution errors}, where jointly processing multiple images causes feature bleeding across the two candidates.

To mitigate these issues, we adopt a decomposed two-stage protocol that analyzes each sub-trajectory independently first, then makes a CoT-style comparison and encourages conservative decisions. 
Notably, this protocol does not \emph{cause} abstention---uncertainty and abstentions are common for VLM oracles when visual evidence is insufficient---but makes it explicit under a conservative prompt, while substantially improving label quality and enabling successful task learning where the original prompt fails.
% \vspace{-5mm}

% =========================================================
% Block B: Adaptive Reward Transition
% =========================================================
\subsection{Bootstrapped Exploration through Adaptive Reward Transition}
\label{sec:exploration}

Even with efficient query selection, preference learning faces a bootstrapping challenge:
meaningful preferences require trajectories with varied task progress,
yet generating such trajectories requires reward guidance that does not initially exist.
Random initial policies often yield aimless motions,
so comparing early trajectories provides little learning signal regardless of query efficiency.

We resolve this circular dependency through an \emph{adaptive reward transition}:
we use VLM-synthesized warm-up rewards to bias early exploration toward task-relevant behaviors,
and then transition to the learned preference-based reward $r_\theta$ once exploration saturates.
This two-phase procedure ensures the policy discovers comparable behaviors worth querying,
while ultimately learning from genuine preferences rather than approximations.

% [EDIT + ADD] Make the legality boundary explicit and crisp.
\subsubsection{Warm-up Reward Synthesis}
We prompt the VLM with the task description $\mathcal{T}_{\text{task}}$ and the environment interface $\mathcal{E}$
to synthesize an executable warm-up reward function
$r_{\text{warm-up}} = \text{VLM-Synthesize}(\mathcal{T}_{\text{task}}, \mathcal{E})$,
following Text2Reward \cite{xie2023text2reward}.
To ensure feasibility for real-world deployment, we allow synthetic rewards to use only realistically measurable geometric cues, such as distances between objects, while prohibiting simulator-privileged signals, e.g., goal coordinates provided by the simulator.
Therefore, this warm-up reward is \emph{not} intended to precisely represent task completion; rather, it aims to encourage coarse task-related progress (e.g., approach, alignment, and initial contact) that enables the collection of informative trajectory pairs for preference learning.

% [EDIT] Trigger: plateau on warm-up return (not success-flag dependent).
\subsubsection{Adaptive Reward Transition}
The policy initially trains under $r_{\text{warm-up}}$ while collecting preference queries in parallel.
We monitor exploration saturation using a generic plateau criterion:
when the warm-up return shows no improvement for $K_{\text{trans}}$ consecutive episodes,
we transition to the now-trained preference-based reward $r_\theta$ for further refinement.

% [KEEP + ADD] Relay exploration: define n_success / schedule succinctly.
\subsubsection{Relay Exploration for Coverage}
To prevent premature collapse to unduly narrow behaviors after transition, 
we implement relay exploration \cite{fu2024furl}:
episodes alternate between the learned policy $\pi$ and a random policy $\pi_{\text{rand}}$ for sampled durations.
We use an inverse-square-root decay schedule $p_{\text{relay}} \propto 1/\sqrt{n_{\text{epi}}}$
for the relay probability,
where $n_{\text{epi}}$ denotes the episode index.
This provides substantial exploratory coverage after transition while progressively reducing relay intervention as learning proceeds.

\section{Experiments}
\label{sec:experiments}

% =========================
% 4.1 Experimental Setup
% =========================
\subsection{Experimental Setup}
\label{sec:setup}

% (tables unchanged; kept as-is for now)
\begin{table*}[t]
  % \centering
  % \small
    \caption{\textbf{Success rates (mean $\pm$ std) across tasks.}
    Middle = 45\% - 55\% steps.
    }
    \label{tab:wide}
      \resizebox{\linewidth}{!}{\begin{tabular}{cccccccc}
\toprule
Method & \textit{Drawer-Open} & \textit{Drawer-Close} & \textit{Button-Press-Topdown} & \textit{Soccer} & \textit{Door-Close} & \textit{Window-Open} & Overall \\
\midrule
VLM-RM \cite{rocamonde2023vision}  & 0.0$\pm$0.0\% & 0.2$\pm$0.5\% & 0.0$\pm$0.0\% & 15.7$\pm$15.3\% & 7.5$\pm$10.5\% & 29.8$\pm$25.9\% & 8.9$\pm$8.7\% \\
Text2Reward \cite{xie2023text2reward} & 17.0$\pm$30.4\% & 79.2$\pm$6.4\% & 22.3$\pm$17.7\% & 6.5$\pm$5.9\% & 99.5$\pm$0.9\% & 54.1$\pm$24.0\% & 46.4$\pm$14.2\% \\
Pebble \cite{lee2021pebble} & 1.5$\pm$2.0\% & 87.1$\pm$5.5\% & 40.6$\pm$9.5\% & 70.5$\pm$26.1\% & 91.0$\pm$6.8\% & 7.8$\pm$5.5\% & 49.8$\pm$9.2\% \\
RL-VLM-F \cite{wang2024rl} & 11.3$\pm$11.2\% & 98.2$\pm$3.6\% & 12.9$\pm$13.4\% & 88.2$\pm$7.9\% & 100.0$\pm$0.0\% & 81.1$\pm$2.7\% & 65.3$\pm$6.5\% \\
\textbf{SPEAR (ours)} &  \textbf{87.0$\pm$15.9\%} &  \textbf{100.0$\pm$0.0\%} &  \textbf{58.2$\pm$16.0\%} &  \textbf{89.9$\pm$11.7\%} &  \textbf{100.0$\pm$0.0\%} &  \textbf{94.3$\pm$4.3\%} &  \textbf{88.2$\pm$8.0\%} \\
\bottomrule
\end{tabular}

}
    \vspace{-0.1in}
\end{table*}

\begin{table*}[t]
  % \centering
  % \small
    \caption{\textbf{Success rates (mean $\pm$ std) across tasks.}
    Final = last 10\% steps.
    }
    \label{tab:wide2}
      \resizebox{\linewidth}{!}{\begin{tabular}{cccccccc}
\toprule
Method & \textit{Drawer-Open} & \textit{Drawer-Close} & \textit{Button-Press-Topdown} & \textit{Soccer} & \textit{Door-Close} & \textit{Window-Open} & Overall \\
\midrule
VLM-RM \cite{rocamonde2023vision} & 0.0$\pm$0.0\% & 0.0$\pm$0.0\% & 0.0$\pm$0.0\% & 14.1$\pm$22.3\% & 3.7$\pm$5.1\% & 71.6$\pm$14.1\% & 14.9$\pm$6.9\% \\
Text2Reward \cite{xie2023text2reward} & 34.8$\pm$32.9\% & 95.8$\pm$4.8\% & 53.7$\pm$27.5\% & 8.9$\pm$5.4\% & 99.8$\pm$0.5\% &  \textbf{87.7$\pm$12.7\%} & 63.4$\pm$14.0\% \\
Pebble \cite{lee2021pebble} & 10.0$\pm$11.6\% & 88.1$\pm$3.2\% & 41.2$\pm$7.7\% & 91.0$\pm$13.3\% & 73.0$\pm$11.5\% & 5.3$\pm$4.1\% & 51.4$\pm$8.6\% \\
RL-VLM-F \cite{wang2024rl} & 33.9$\pm$21.4\% & 98.8$\pm$3.1\% & 17.9$\pm$16.6\% & 75.4$\pm$11.6\% & 95.7$\pm$7.6\% & 69.2$\pm$5.3\% & 65.1$\pm$10.9\% \\
% \textbf{VL-PEARL} & 86.4$\pm$12.7\% & 100.0$\pm$0.0\% & 67.0$\pm$17.8\% & 95.1$\pm$3.9\% & 100.0$\pm$0.0\% & 84.9$\pm$5.7\% & 88.9$\pm$6.7\% \\
\textbf{SPEAR (ours)} & \textbf{86.4$\pm$12.7\%} & \textbf{100.0$\pm$0.0\%} & \textbf{67.0$\pm$17.8\%} & \textbf{95.1$\pm$3.9\%} & \textbf{100.0$\pm$0.0\%} & 84.9$\pm$5.7\% & \textbf{88.9$\pm$6.7\%} \\
\bottomrule
\end{tabular}}
    \vspace{-0.1in}
\end{table*}

% [UNCHANGED] (kept verbatim)
% \noindent\textbf{Tasks and environment.}
\subsubsection{Tasks and Environment}
We evaluate \method on six manipulation tasks
from the Meta-World benchmark~\cite{yu2020meta}:
\textit{drawer-open}, \textit{drawer-close}, \textit{button-press-topdown},
\textit{soccer}, \textit{door-close}, and \textit{window-open}.
% These tasks exhibit diverse challenges, including contact-rich manipulation (\textit{drawer-open}, \textit{window-open}),
% precise positioning (\textit{button-press-topdown}), and dynamic object interactions (\textit{soccer}).
All tasks use image observations for VLM queries and reward model training,
with RL policies trained via Soft Actor-Critic (SAC)~\cite{haarnoja2018soft}.

% [EDIT] Baselines
%   - Deleted: "which performs similarly to GPT-4o~\cite{hurst2024gpt}."
%   - Added: (i) same open-source oracle for fairness, (ii) diagnostics averaged over 5 backbones (Sec. 4.2.3).
% \noindent\textbf{Baselines.}
\subsubsection{Baselines}
We compare against four representative VLM-guided RL approaches:
\textbf{(1) VLM-RM}~\cite{rocamonde2023vision}
constructs rewards through cosine similarity between BLIP2 embeddings and task descriptions;
\textbf{(2) Text2Reward}~\cite{xie2023text2reward}
leverages VLMs to generate executable reward functions;
\textbf{(3) PEBBLE}~\cite{lee2021pebble} implements PbRL with
entropy-based query selection and a scripted oracle
(Meta-World's ground-truth rewards);
\textbf{(4) RL-VLM-F}~\cite{wang2024rl}
applies VLM preferences with random query selection.
For fair comparison, we employ the same open-source preference oracle, InternVL3-78B~\cite{zhu2025internvl3}, and set sub-trajectory length $H=1$ for both RL-VLM-F and our method.
In addition, we report VLM reliability diagnostics averaged over five VLM backbones in Sec.~\ref{sec:oracle_diagnostics} (including Qwen3-VL and InternVL3.5 variants~\cite{yang2025qwen3,wang2025internvl3.5,zhu2025internvl3}).

% [NEW][Issue B,F] Oracle/VLM query protocol & abstention (descriptive only; no raw prompts in main text)
% \noindent\textbf{Oracle query protocol and abstention.}
% \subsubsection{Oracle query protocol and abstention}
% Each oracle query compares two observations and returns either a preference (which image is closer to completing the task) or abstains when a reliable decision cannot be made.
% To mitigate ``confident-but-wrong'' feedback induced by naive joint-image prompting, we adopt a decomposed query protocol during training:
% the oracle first analyzes each image independently, then performs a chain-of-thought style comparison and outputs a final preference or abstention.
% We use a conservative preference policy in training (abstain when not confident), while a more radical policy is used only in offline diagnostics (Sec.~\ref{sec:oracle_diagnostics}).

% [EDIT] Evaluation protocol: add fixed query cadence + no extra re-query on abstain; keep rest concise.
% \noindent\textbf{Evaluation protocol.}
\subsubsection{Evaluation Schemes}
\label{sec:setup_eval}
Each method is evaluated across 5 random seeds
with mean success rates and standard deviations reported.
Policies undergo evaluation every 10K
environment steps over 50 episodes.
Oracle queries are scheduled with a fixed step-coupled cadence:
every 10K environment steps, we collect 40 preference comparisons and update the reward model (and our distinguishability predictor).
Abstentions do not trigger additional queries; therefore, the total query budget is fixed by the training steps under the same schedule.
All experiments utilize 8 NVIDIA A800 GPUs.

% [NEW][Issue A,F] Cost metrics & wall-clock measurement (short "statute" paragraph for 4.2 tables)
% \noindent\textbf{Cost metrics and wall-clock measurement.}
% \subsubsection{Cost Metrics and Wall-clock Measurement}
% We report total oracle queries ($Q$), abstention rate ($\rho$), and effective queries $Q_{\text{eff}} = Q(1-\rho)$.
% We additionally report the average oracle latency per comparison (AvgQ, seconds).
% To isolate the method-specific overhead beyond standard RL training (shared across methods), we summarize (i) total oracle inference time and (ii) the average per-round training time of the distinguishability predictor (seconds per 10K-step round), and compare the resulting extra wall-clock cost at a matched success-rate level set.

\begin{figure*}[t]
  \centering
  \includegraphics[width=0.86\textwidth]{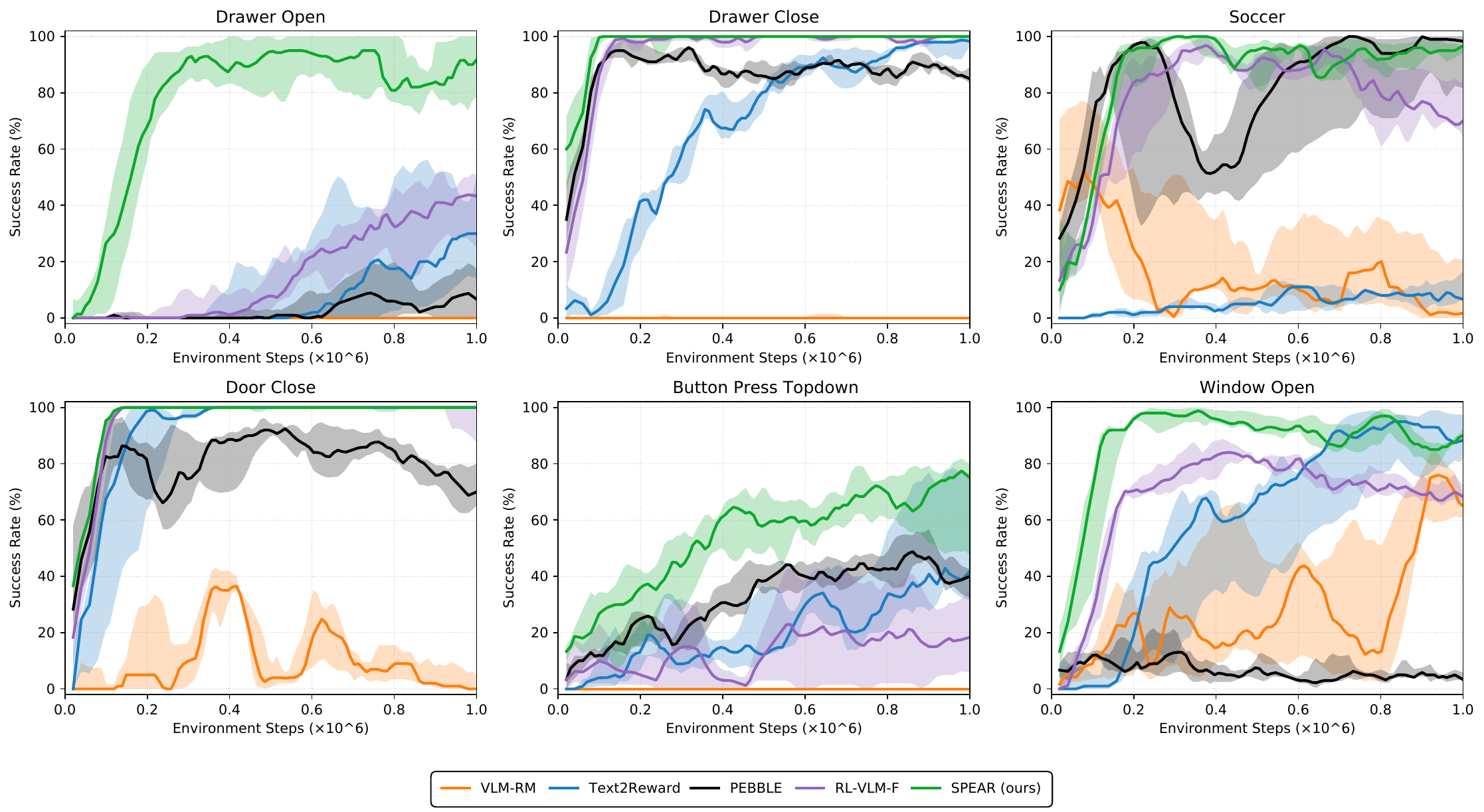}
  \caption{
    \textbf{Learning curves across six Meta-World tasks.} 
    Our method (green) consistently outperforms baselines in both mid-range and final performance. Shaded regions indicate standard deviation over 5 seeds.
  }
  \vspace{-3mm}
  \label{fig:main_curves}
\end{figure*}

% =========================
% 4.2 Main Results (Simulation only)
% =========================
\subsection{Main Results}
\label{sec:main_results}

\subsubsection{Task Performance}
\label{sec:task_performance}

% [MINOR EDIT] Kept largely verbatim; only minimal tightening/grammar where necessary.
Tables~\ref{tab:wide} and~\ref{tab:wide2} present success rates at mid-training and final evaluation, respectively, while Figure~\ref{fig:main_curves} illustrates learning curves.
\method~demonstrates consistent superiority across tasks of varying difficulty,
achieving an overall success rate of 88.2\% at mid-training and 88.9\% at final evaluation -- outperforming the RL-VLM-F baseline by a margin of 23.3\%.

The performance gap is particularly significant in contact-rich tasks requiring precise reward shaping, such as \textit{drawer-open} and \textit{button-press-topdown}. This indicates that our distinguishability-aware query selection and reward transition enable the reward model to capture task progress more promptly than random sampling, and when equipped with efficient query selection, \method can refine reward signals to a level far exceeding that provided by the coarse VLM synthesis function.

On \textit{window-open} task, Text2Reward achieves a slightly higher final success rate (87.7\% versus 84.9\%). Analysis reveals this task's reward structure aligns particularly well with VLM-synthesized distance-based objectives, but considering the overall task performance, the synthesized reward is insufficient to support task completion stably and reliably. This suggests that our preference-based approach has a wider applicability and can be applied to various task semantics. 
% In summary, \method maintains excellent overall performance while significantly reducing training costs.

\subsubsection{Query Efficiency Analysis}
\label{sec:query_efficiency}

% [EDIT] Rewritten to (i) make cadence->derived queries explicit, (ii) generalize beyond drawer-open, (iii) isolate extra-time accounting.
Beyond task success, we evaluate the practical efficiency of online training.
Because oracle queries are scheduled at a fixed cadence (Sec.~\ref{sec:setup_eval}), the total query count is proportional to environment steps; thus, improved success rate at the same number of environment steps directly implies improved query-efficiency.
We choose \textit{drawer-open} and compare \method\ against RL-VLM-F at a matched 30\% success-rate level in the setting without reward transition, see Table~\ref{tab:per_task_efficiency}.
\method~uses 960 total queries with 29.6\% abstention, compared to RL-VLM-F's 1640 queries with 55.5\% abstention, resulting in substantially fewer wasted comparisons while achieving faster learning.
In terms of wall-clock, although \method\ incurs additional overhead to train the distinguishability predictor at each query round, the total time to reach 30\% success is reduced from 13.6 hours to 9.1 hours.
This is because the savings from avoiding wasted queries more than offset the overhead of distinguishability-model training.

% [NEW][Issue B,C] Oracle reliability diagnostics (averaged over 5 backbones; offline only).
\subsubsection{VLM Reliability Diagnostics}
\label{sec:oracle_diagnostics}
A key concern is whether VLM feedback can exhibit a ``confident but incorrect'' behavior when the VLM does not abstain.
We therefore run a controlled diagnostic study on five VLM backbones (Qwen3-VL-8B/30B-A3B, InternVL3.5-8B/30B-A3B, and InternVL3-78B~\cite{yang2025qwen3,wang2025internvl3.5,zhu2025internvl3}) to quantify abstention versus correctness under different visual distinguishability. 

For each task, we vary visual distinguishability by altering the frame gap in a single demo,
and test both \textit{radical} and \textit{conservative} prompts (adding a style suffix to the original prompt).
Importantly, these demos are used only for offline analysis and never for online training or query collection.

We find that our decomposed protocol markedly eliminates the systematic positional bias and feature bleeding caused by simple joint image pairwise cues.
Table~\ref{tab:conservative_abs_frame_diff} shows that the abstention rate consistently decreases with increasing visual distinguishability, while accuracy steadily improves, reaching 100\% at sufficiently large frame gaps.
Under the \textit{conservative} prompting style, the average preference accuracy of the five backbones reached 92.3\%, 
demonstrating that VLM preference feedback is reliably accurate in our setup with an appropriate query formula.

% [NEW][Issue A,F] Per-task (derived) query efficiency at a matched success-rate level set.
% NOTE: Use this as a template; you will replace TBD values with actual measurements.
% \begin{table*}[t]
%   \centering
%   \caption{\textbf{Query efficiency and wall-clock cost at 30\% success.}
%   \method effectively reduces the abstention rate and lowers the training wall-clock time overhead.
%   }
%   \label{tab:per_task_efficiencyaa}
%   \resizebox{\linewidth}{!}{%
%     \begin{tabular}{llrrrrrrrr}
%       \toprule
%       Method &
%       Steps@30\% (k) &
%       TotalQ &
%       Abst.(\%) &
%       EffQ &
%       AvgQ (s) &
%       OracleTime (h) &
%       DistTrain (s/round) &
%       ExtraTime (h) \\
%       \midrule
%       RL-VLM-F & 410 & 1640 & 55.5 & 729 & 29.87 & 13.6 & -- & 13.6 \\
%       \textbf{\method} & 240 & 960 & 29.6 & 676 & 31.42 & 8.4 & \textit{TBD} & \textit{TBD} \\
%       \bottomrule
%     \end{tabular}%
%   }
% \end{table*}

\begin{table}[t]
  \centering
  \caption{\textbf{Query efficiency and wall-clock cost at 30\% success.}  
  \method effectively reduces the abstention rate and lowers the training wall-clock time overhead. 
  }
  \label{tab:per_task_efficiency}
  \resizebox{\columnwidth}{!}{%
    \begin{tabular}{lcccccc}
      \toprule
      \textbf{Method} &
      % \multirow{2}{*}{\textbf{Method}} &
      \multicolumn{3}{c}{\textbf{Query Efficiency}} &
      \multicolumn{3}{c}{\textbf{Time Consumption}} \\
      \cmidrule(lr){2-4} \cmidrule(lr){5-7}
      & Total Q & Abst. Rate & Effective Q & Aver. Q (s) & DistTrain (s) & Total (h)\\
      \midrule
      RL-VLM-F
      & 1640 & 0.555 & 729 & 29.87 & -- & 13.6 \\
      \textbf{\method~(ours)}
      & 960 & 0.296 & 676 & 31.42 & 114.5 & 9.1 \\
      \bottomrule
    \end{tabular}%
  }
  \vspace{-0.1in}
\end{table}

\begin{table}[t]
\caption{
    \textbf{Diagnostics of averaged 5 VLM backbones with \textit{conservative} prompt.} Distinctiveness increases $\Rightarrow$ abstention decreases and accuracy increases.
} 
\label{tab:conservative_abs_frame_diff}
\centering
% \small
\renewcommand{\arraystretch}{0.5}
\resizebox{\linewidth}{!}{%
\begin{tabular}{llccccc}
\toprule
Frame Gap & Metric (\%) & \textit{Drawer-Open} & \textit{Door-Close} & \textit{Soccer} & \textit{Button-Press} & Avg \\
\midrule
\multirow{2}{*}{1--5} & Abstain & 61.2 & 88.8 & 77.2 & 76.6 & 75.9 \\
 & Acc & 61.3 & 75.0 & 53.7 & 56.0 & 61.5 \\
\midrule
\multirow{2}{*}{6--10} & Abstain & 61.6 & 87.5 & 64.6 & 68.1 & 70.5 \\
 & Acc & 79.6 & 100.0 & 91.3 & 68.6 & 84.9 \\
\midrule
\multirow{2}{*}{11--20} & Abstain & 49.2 & 82.0 & 72.7 & 65.6 & 67.4 \\
 & Acc & 91.3 & 100.0 & 100.0 & 97.0 & 97.1 \\
\midrule
\multirow{2}{*}{21--30} & Abstain & 29.5 & 60.7 & -- & 54.7 & 48.3 \\
 & Acc & 100.0 & 100.0 & -- & 97.7 & 99.2 \\
\midrule
\multirow{2}{*}{31--max} & Abstain & 11.6 & -- & -- & 45.0 & 28.3 \\
 & Acc & 99.4 & -- & -- & 100.0 & 99.7 \\
\bottomrule
\end{tabular}%
}
\vspace{-0.1in}
\end{table}

% =========================
% 4.3 Ablation Studies
% =========================
\subsection{Ablation Studies}
\label{sec:ablations}

% [EDIT] Expanded intro to (i) specify task scope, (ii) state simulated-oracle usage for all ablations.

% We conduct systematic ablations to isolate and validate the contributions of \method's core mechanisms.

% All ablations in this section are conducted on \textit{drawer-open} (the most contact-rich and reward-shaping-sensitive task in our benchmark) and use a calibrated simulated preference oracle described below to enable extensive controlled comparisons.

% [NEW][Issue G] Validity note for simulated oracle used in ablations (clearer + includes GLM-GAM(B-spline) + approximation boundaries).
\subsubsection{Validity of Simulated Preference Oracle}
\begin{figure}[t]
  \centering
  \includegraphics[width=0.85\linewidth]{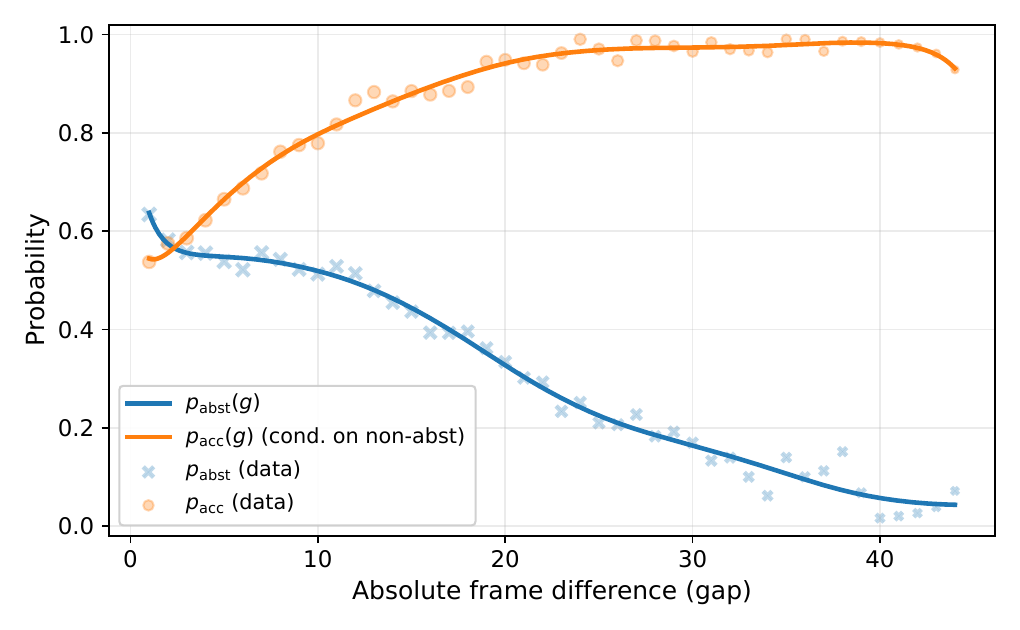}
  \caption{
    \textbf{Average fit probability of 5 VLM backbone behaviors.}
    As visual distinguishability increases, the abstention rate decreases and the accuracy increases.
  }
  \vspace{-0.2in}
  \label{fig:gap_fit}
\end{figure}

To run large-scale ablations without repeatedly invoking expensive VLM inference, we use a calibrated \emph{simulated} preference oracle that emulates the aggregate behavior of five open-source VLM backbones (as diagnosed in Sec.~\ref{sec:oracle_diagnostics}).
We model the oracle as a stochastic labeler:
it abstains with probability $p_{\text{abst}}(\cdot)$; otherwise it outputs the correct preference label with probability $p_{\text{acc}}(\cdot)$ or flips it with probability $1-p_{\text{acc}}(\cdot)$.
We fit smooth functions $p_{\text{abst}}(g)$ and $p_{\text{acc}}(g)$ from aggregated diagnostic statistics via a binomial Generalized Additive Model (GAM) with B-spline smooths, as illustrated in Figure~\ref{fig:gap_fit}.
During training, we use the return difference $\Delta R$ from simulation reward between two sub-trajectories as a practical proxy for distinguishability and linearly map $\Delta R$ to $g$ using a per-task reward range estimated from demonstrations.
While viewing $\Delta R$ as visual distinguishability is heuristic, this stochastic labeler accurately aligns with the monotonic trend of VLM behavior in  Sec.~\ref{sec:oracle_diagnostics}. This simulated oracle is used \emph{only} in the ablation section.

\subsubsection{Query Acquisition Mechanisms}
\label{sec:ablation_query}

% [EDIT] Minimal edits: fix typos, tighten phrasing, align with simulated-oracle context (no "perfect labeling" assumption claims beyond oracle behavior).
Table~\ref{tab:ablation_query} compares four query selection strategies on \textit{drawer-open}.
Random selection yields only 14.0\% final success, confirming that uninformed sampling provides insufficient learning signal.
Entropy-only (standard BALD) improves marginally to 21.0\% but often selects pairs that are highly uncertain yet practically unanswerable (high abstention probability).

Distinguishability-only selection achieves 69.7\% success by avoiding abstentions, 
but plateaus as it favors easily distinguishable pairs that may lack information gain.
Our integrated approach synthesizes both criteria, reaching 80.7\% final success with 58.4\% at mid-training.
This validates that jointly optimizing answerability and informativeness is essential for efficient preference-based learning.

% [UNCHANGED] Table input kept as-is (you will polish labels later).
\begin{table}[t]
\centering
\small
\caption{
    \textbf{Ablation on query acquisition mechanisms.}
    Integrated weighting (ours) achieves the best performance.
}
  \label{tab:ablation_query}
     \resizebox{\linewidth}{!}{\begin{tabular}{lcc}
\toprule
\textbf{Method} & \textbf{Mid (450k–550k)} & \textbf{End (900k–1000k)} \\
\midrule
Random & 1.1 $\pm$ 1.8 \% & 14.0 $\pm$ 15.9 \% \\
Entropy-only & 3.3 $\pm$ 4.2 \% & 21.0 $\pm$ 19.0 \% \\
Distinguishability-only & 21.9 $\pm$ 19.8 \% & 69.7 $\pm$ 24.3 \% \\
\textbf{Integrated (ours)} & \textbf{58.4 $\pm$ 23.8 \%} & \textbf{80.7 $\pm$ 13.2 \%} \\
\bottomrule
\end{tabular}}
\vspace{-0.1in}
\end{table}

% [UNCHANGED] Exploration table input kept as-is.
\begin{table}[t]
\centering
\small
\caption{
    \textbf{Ablation on exploration mechanisms.}
    Combining two mechanisms achieves the best performance.
  }
  \label{tab:ablation_exploration}
     \resizebox{\linewidth}{!}{\begin{tabular}{lcc}
\toprule
\textbf{Method} & \textbf{Mid (450k–550k)} & \textbf{End (900k–1000k)} \\
\midrule
Baseline & 1.7 $\pm$ 2.5 \% & 24.8 $\pm$ 16.2 \%  \\
+ Relay Exploration & 26.1 $\pm$ 21.6 \% & 73.6 $\pm$ 19.5 \% \\
+ Reward Transition & 39.8 $\pm$ 17.8 \% & 67.0 $\pm$ 21.9 \% \\
\textbf{Full Exploration (ours)} & \textbf{92.7 $\pm$ 8.9 \%} & \textbf{88.3 $\pm$ 11.7 \%} \\
\bottomrule
\end{tabular}

  % \begin{tabular}{lc}
  %   \toprule
  %   \textbf{Configuration} & \textbf{Success Rate (\%)} \\
  %   \midrule
  %   Baseline (random init + $\epsilon$-greedy) & [XX.X$\pm$XX.X] \\
  %   + Reward Switching                          & [XX.X$\pm$XX.X] \\
  %   + Relay Exploration                         & [XX.X$\pm$XX.X] \\
  %   \midrule
  %   \textbf{Full Method (Ours)}                 & \textbf{[XX.X$\pm$XX.X]} \\
  %   \bottomrule
  % \end{tabular}

%   \begin{tabular}{lll}
% \toprule
%  & Mid (450k--550k) & End (900k--1000k) \\
% \midrule
% baseline & 1.7 $\pm$ 2.5 \% & 24.8 $\pm$ 16.2 \% \\
% + relay exploration & 26.1 $\pm$ 21.6 \% & 73.6 $\pm$ 19.5 \% \\
% + reward switching & 39.8 $\pm$ 17.8 \% & 67.0 $\pm$ 21.9 \% \\
% full exploration (ours) & 92.7 $\pm$ 8.9 & 88.3 $\pm$ 11.7 \% \\
% \bottomrule
% \end{tabular}
}
\vspace{-0.1in}
\end{table}

% [NEW] Ultra-compact integration of extended analysis (no extra tables): distinguishability predictor validation + calibration.
\subsubsection{Distinguishability Model Validation}

We additionally isolate and evaluate the distinguishability predictor $\mathcal{D}_\psi$ using oracle-response data collected from \textit{drawer-open} to validate its predictive accuracy. We report the model’s Mean Squared Error (MSE, i.e., Brier score) and Mean Absolute Error (MAE) to evaluate the model’s predictive performance.

% 可选：全局压缩环绕对象与文字间距
% \setlength{\intextsep}{8pt}   % 环绕对象上下与正文的间距
% \setlength{\columnsep}{10pt}  % 双栏间距（一般别乱动，必要时再改）
\begin{wraptable}{r}{0.52\columnwidth}
% \vspace{-6pt} % 视情况调：把表往上提一点
\centering
\caption{\textbf{Feature/calibration ablation.} Multi-scale features and temperature scaling improve prediction reliability.}
\label{tab:feature_ablation}
\footnotesize
\setlength{\tabcolsep}{3pt}
\renewcommand{\arraystretch}{0.95}
\begin{tabular}{@{}lcc@{}}
\toprule
Configuration & MSE$\downarrow$ & MAE$\downarrow$ \\
\midrule
Global only ($\phi_{\text{rel}}$) & 0.1542 & 0.2973 \\
Local only ($\phi_{\text{loc}}$) & 0.1662 & 0.3380 \\
\textbf{Combined (ours)} & \textbf{0.1508} & \textbf{0.2950} \\
\midrule
w/o calibration & 0.1610 & 0.3232 \\
\textbf{w/ calibration (ours)} & \textbf{0.1515} & \textbf{0.2978} \\
\bottomrule
\end{tabular}
% \vspace{-8pt} % 视情况调：压缩表下方空白
\end{wraptable}

% \begin{table}[t]
% \centering
% \caption{\textbf{Feature and calibration ablation on distinguishability prediction.} Multi-scale features achieve the lowest prediction error, and temperature scaling improves prediction reliability.}
% \label{tab:feature_ablation}
% % \small
% \setlength{\tabcolsep}{4pt}
% \begin{tabular}{@{}lcc@{}}
% \toprule
% \textbf{Configuration} & \textbf{MSE} $\downarrow$ & \textbf{MAE} $\downarrow$ \\
% \midrule
% Global only ($\boldsymbol{\phi}_{\text{rel}}$) & 0.1542 & 0.2973 \\
% Local only ($\boldsymbol{\phi}_{\text{loc}}$) & 0.1662 & 0.3380 \\
% \textbf{Combined (ours)} & \textbf{0.1508} & \textbf{0.2950} \\
% \midrule
% w/o calibration & 0.1610 & 0.3232 \\
% \textbf{w/ calibration (ours)} & \textbf{0.1515} & \textbf{0.2978} \\
% \bottomrule
% \end{tabular}
% \end{table}

Table~\ref{tab:feature_ablation} shows that using multi-scale features achieves the lowest prediction error (Brier score 0.1508, MAE 0.2950), validating that the synergy between global semantic understanding and local spatial discrimination is critical, while applying temperature scaling improves calibration quality and reduces both MSE and MAE. The combined effect of both directly translates to more effective query selection and a lower abstention rate during online training.

% [EDIT] Keep your original logic; only minor tightening and a clean ending.
\subsubsection{Exploration Mechanisms}
\label{sec:ablation_exploration}

Table~\ref{tab:ablation_exploration} exhibits our two-phase exploration strategy.
The baseline configuration with random initialization and standard $\epsilon$-greedy exploration achieves only 25.6\% final success, illustrating the bootstrapping challenge.
Adding reward transition alone improves mid-training performance to 39.8\%, confirming that semantic bootstrapping jumpstarts exploration toward task-relevant regions.
However, 
% $r_{\text{warm-up}}$ provides limited pre-guidance; 
after the transition, the agent still faces the challenge of efficiently discovering better states beyond the guided regime.

Relay exploration alone maintains distributional coverage, but it cannot replace the directional guidance from warm-up rewards.
The full method combining both mechanisms achieves 92.7\% mid-training and 88.3\% final success.
This synergy emerges because $r_{\text{warm-up}}$ enables initial progress while relay exploration prevents the policy from collapsing to narrow behaviors,
ensuring sufficient trajectory diversity for meaningful preference comparisons.
% Overall, these ablations validate that practical preference-based RL with imperfect oracles requires addressing query efficiency and exploration bootstrapping as interdependent challenges.

% =========================
% 4.4 Real-Robot Experiments
% =========================
% [NEW] Entire section.
\subsection{Real-Robot Experiments}
\label{sec:real_robot}

\begin{figure}[t]
  \centering
  \includegraphics[width=\linewidth]{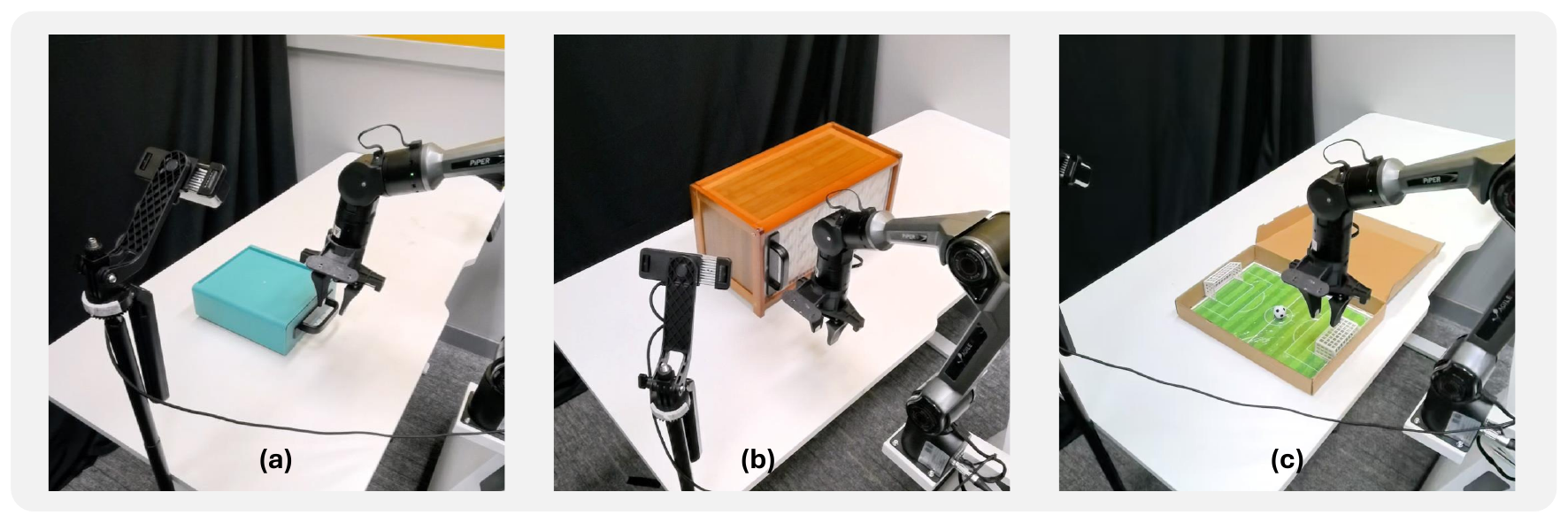}
  \caption{
    \textbf{Real-robot experiment task settings.} 
    (a) \textit{drawer-open}. (b) \textit{window-open}. (c) \textit{soccer}.
  }
  \vspace{-3mm}
  \label{fig:real_robot}
\end{figure}

% [NEW] High-level takeaway: real-robot online training, no sim2real.
We evaluate \method\ on a real Mobile ALOHA robot, running the full online exploration and preference-based learning pipeline directly in the physical environment.
Therefore, our experiment does not involve sim-to-real transfer.

% [NEW] Platform + sensing + action space.
\subsubsection{Platform and Tasks}
We use the left arm of Mobile ALOHA with a fixed side-view RGB camera.
The action space matches our simulator setting: Cartesian end-effector increments in $(x,y,z)$ and a gripper open/close command.
We evaluate three manipulation tasks: \textit{drawer-open}, \textit{window-open}, and \textit{soccer} (ball pushing), as shown in Figure~\ref{fig:real_robot}.

% [NEW] Online training protocol, cadence, and oracle choice.
\subsubsection{Online training protocol}
Each episode consists of 50 exploration steps.
At each step, the end-effector motion is bounded by a maximum displacement of 2\,cm, executing a step takes approximately 1\,s, and we record the current RGB observation. At the end of each episode, the arm returns to an initial pose and waits for 5\,s before the next episode begins. 
As a practical constraint of real-robot deployment, we manually reset the object to the initial configuration between episodes.

Every 10 episodes (500 steps), we enter the preference-query branch and request 20 preference comparisons to update the reward model (and the distinguishability model) following the same overall pipeline as Sec.~\ref{sec:setup}. We use Qwen3-VL-8B-Instruct \cite{yang2025qwen3} as the preference oracle, running on a local GPU workstation. 

% We periodically evaluate the latest checkpoint in the real environment. The task is declared success when the policy successfully achieves 5 consecutive evaluation rollouts. We run one trial per task and report the steps/query budget required to achieve stable success.

% [NEW] Summary table (leave room for later polishing).
% \begin{table}[t]
% \centering
% \caption{\textbf{Real-robot online training on Mobile ALOHA.}
% Steps report the first checkpoint that achieves 5 consecutive successful evaluations.
% Query rounds occur every 10 episodes (20 comparisons each).
% Robot time is approximate and excludes manual environment resets and oracle inference overhead.}
% \label{tab:real_robot}
% \small
% \begin{tabular}{lrrrrr}
% \toprule
% Task & Steps & Episodes & QueryRounds & Comparisons & RobotTime (h) \\
% \midrule
% drawer-open & 18200 & 364 & 36 & 720 & 5.56 \\
% window-open & 16600 & 332 & 33 & 660 & 5.07 \\
% soccer      & 12600 & 252 & 25 & 500 & 3.85 \\
% \bottomrule
% \end{tabular}
% \end{table}

% [NEW] Results narrative (kept short, non-overclaiming).
\subsubsection{Results}
\method\ learns all three tasks directly on the physical robot with end-to-end online interaction.
The policy converges in 18.2K steps for \textit{drawer-open}, 16.6K steps for \textit{window-open}, and 12.6K steps for \textit{soccer}.
Under our fixed real-robot query cadence, this corresponds to 720, 660, and 500 preference comparisons, respectively, demonstrating that the overall pipeline remains practical under real-time constraints.

% [NEW] Safety / failure modes + mitigation (explicit but concise).
\subsubsection{Safety and failure modes}
A primary failure mode in unconstrained exploration is unsafe contact, where the robotic arm may apply excessive force to an object (e.g., pushing an object's handle forcefully).
To mitigate this, we (i) restrict the exploration workspace to a safe operating region while preserving sufficient coverage for learning, and (ii) add a small penalty when the commanded end-effector displacement cannot be realized due to collisions or environmental constraints, discouraging repeated unsafe actions. Overall, the experiments demonstrate that \method can be deployed on real hardware and in real-world environments as a fully online learning system.

\section{Conclusion}
\label{sec:conclusion}

We presented \method, 
a practical framework 
for learning robotic manipulation 
via VLM-guided preference-based reinforcement learning. 
By addressing query efficiency 
and exploration bootstrapping 
as coupled challenges, 
our approach enables 
effective online learning despite the high cost of VLM queries.
% Extensive experiments of the synergistic interaction 
% between distinguishability-aware acquisition 
% and adaptive reward transition 
Experiments demonstrate that 
% practical deployment 
% requires co-designing system components 
% rather than optimizing individual modules. 
% Extensive experiments show 
\method~significantly outperforms existing methods 
in both sample efficiency and final performance 
across diverse tasks, 
establishing a foundation for 
scalable robot learning 
without manual reward engineering 
or human annotation.

\section*{ACKNOWLEDGMENT}

This work was supported by the Early Career Scheme of the Research Grants Council (RGC) under Grant \# 27207224, the HKU-100 Award, a donation from the Musketeers Foundation, and in part by the JC STEM Lab of Autonomous Intelligent Systems funded by The Hong Kong Jockey Club Charities Trust.

% =======================
% References
% =======================
\bibliographystyle{IEEEtran}
\bibliography{ref}

\end{document}